\pdfoutput=1

\documentclass[11pt]{article}

\usepackage[final]{acl}

\usepackage{times}
\usepackage{latexsym}

\usepackage[T1]{fontenc}

\usepackage[utf8]{inputenc}

\usepackage{microtype}

\usepackage{inconsolata}

\usepackage{graphicx}

\usepackage[utf8]{inputenc} 
\usepackage[T1]{fontenc}    
\usepackage{hyperref}
\usepackage{booktabs}       
\usepackage{amsfonts}       
\usepackage{nicefrac}       
\usepackage{microtype}      
\usepackage{xcolor}         
\usepackage{wrapfig}

\usepackage{microtype}
\usepackage{graphicx}
\usepackage{subfigure}
\usepackage{booktabs} 

\usepackage{amsmath}
\usepackage{amssymb}
\usepackage{mathtools}
\usepackage{amsthm}
\usepackage{mathrsfs}
\usepackage{graphicx}

\usepackage[normalem]{ulem}
\useunder{\uline}{\ul}{}
\usepackage{colortbl}
\usepackage{capt-of}
\usepackage[]{graphicx}
\usepackage{pifont}
\usepackage[capitalize,noabbrev]{cleveref}
\usepackage{gensymb}

\usepackage[utf8]{inputenc} 
\usepackage[T1]{fontenc}    
\usepackage{url}            
\usepackage{booktabs}       
\usepackage{amsfonts}       
\usepackage{nicefrac}       
\usepackage{microtype}      
\usepackage{xcolor}         
\definecolor{blue}{rgb}{0,0,1}
\definecolor{mutedblue}{rgb}{0.23, 0.37, 0.80}
\definecolor{darkgreen}{rgb}{0,0.40,0}
\definecolor{firebrick}{rgb}{0.698,0.133,0.133}

\newcommand*{\green}[1]{\textcolor{darkgreen}{#1}}

\hypersetup{
    colorlinks=true,
    linkcolor=firebrick,
    citecolor=mutedblue,
    urlcolor=firebrick,
    linktocpage=true,
    unicode=true,
    pdfborder={0 0 0} 
}

\makeatletter
\renewcommand{\@makefnmark}{\hbox{\textsuperscript{\normalfont\color{black}\@thefnmark}}}
\makeatother

\theoremstyle{plain}

\theoremstyle{definition}

\theoremstyle{remark}

\usepackage{lipsum}
\usepackage{multirow}
\usepackage{booktabs}
\usepackage{hyperref}
\usepackage{url}
\usepackage{float}

\title{Can LLMs Understand Unvoiced Speech?\\Exploring EMG-to-Text Conversion with LLMs}
\author{
Payal Mohapatra\thanks{\tiny Equal contribution}\footnotemark[2] \quad
Akash Pandey\footnotemark[1] \quad
Xiaoyuan Zhang\footnotemark[1] \quad
Qi Zhu\thanks{\tiny Corresponding authors:\texttt{\{payal.mohapatra, qzhu\}@northwestern.edu}} \\
Northwestern University, USA
}

\begin{document}
\maketitle
\begin{abstract}

Unvoiced electromyography (EMG) is an effective communication tool for individuals unable to produce vocal speech. However, most prior methods rely on paired voiced and unvoiced EMG signals, along with speech data, for unvoiced EMG-to-text conversion, which is not practical for these individuals. Given the rise of large language models (LLMs) in speech recognition, we explore their potential to understand unvoiced speech. To this end, we address the challenge of \emph{learning from unvoiced EMG alone} and propose a novel EMG adaptor module that maps EMG features to an LLM’s input space, achieving an average word error rate of 0.49 on a closed-vocabulary unvoiced EMG-to-text task. Even with a conservative data availability of just six minutes, our approach improves performance over specialized models by nearly 20\%. While LLMs have been shown to be extendable to new language modalities—such as audio—understanding articulatory biosignals, like unvoiced EMG, is more challenging. This work takes a crucial first step toward enabling LLMs to comprehend unvoiced speech using surface EMG.

 
\end{abstract}

\section{Introduction and Related Works}\label{sec:intro}


Speech impairments affect around 4 million people in the U.S. alone~\cite{NIDCD2024}. Silent speech interfaces~\cite{zhang2021speechin, mohapatra2024non, gonzalez2020silent,srivastava2024poster} have emerged as transformative solutions, enabling communication for individuals who cannot rely on spoken language. One such instrument is surface electromyography (EMG)~\cite{schultz2017biosignal}, which captures muscle activations crucial for speech production, even during unvoiced articulation. Inspired by the immense success of the speech recognition abilities of text-to-text large language models (LLMs)~\cite{tang2023salmonn, yu2024connecting, leveraging-large-language-models}, we study an important question—\textit{what is the potential of these LLMs to understand unvoiced speech?} Specifically, can they convert silent speech to text without access to audio or voiced versions of the EMG signals? This question is particularly relevant for assisting individuals who can no longer produce audible speech~\cite{meltzner2017silent}, where no corresponding voiced EMG or speech data exists. Additionally, given the highly personal nature of these signals~\cite{diener20_interspeech, wand2009impact}, it is crucial to develop methods that learn effectively from limited unvoiced EMG data.


Prior works~\cite{jou2006towards, meltzner2018development, schultz2010modeling} on EMG-to-text conversion focused primarily on voiced signals. Other studies~\cite{gaddy-klein-2021-improved, gaddy2020digital, benster2024cross} on unvoiced EMG-to-text conversion leveraged auxiliary tasks to align unvoiced EMG with audio from voiced sessions or applied strategic transfer learning from voiced EMG-audio models, both of which rely on vocal data. However, we consider a scenario where no voiced signals are available for a speaker and explore a technique to communicate with LLMs, the modern workhorses for language understanding~\cite{dubey2024llama, openai2023gpt}. Recent research successfully expanded LLMs to other language modalities, such as speech~\cite{tang2023salmonn, yu2024connecting} and silent video~\cite{maaz2023video, yeo-etal-2024-visual}. A key approach involves adaptor modules—ranging from simple trainable linear layers~\cite{ma2024embarrassingly} to dedicated projector networks~\cite{kang2024frozen} and explicit alignment strategies~\cite{li2023blip, tan2024ssr}—to map new language modalities to LLMs’ input embedding space. While \citet{benster2024cross} incorporated LLMs as a post-processing step after multimodal EMG model predictions, it largely remains unexplored whether effective unvoiced EMG-to-text conversion can be achieved by directly leveraging LLMs.






\smallskip
\noindent\textbf{Our Contributions.} We study a practical setting of converting unvoiced EMG to text without access to voiced EMG or audio, by expanding LLMs to understand this new language modality. We propose a novel trainable EMG adaptor module to map EMG features into the LLM's input space. Our approach, focused on a closed vocabulary, demonstrates promising results, achieving an average word error rate of 0.49. With just six minutes of training data, LLMs outperform specialized EMG-to-text models by 20\%. We analyze the EMG adaptor’s design, performance across varying data amounts and features, and broader challenges in learning from unvoiced EMG. Our work paves the way for integrating unvoiced EMG with large language models (LLMs), improving their text conversion accuracy and enabling non-vocal users to fully benefit from LLM-based assistants.



\section{Approach}\label{sec:approach}
\noindent \textbf{Adaptor Network Design.} The input unvoiced EMG signals are represented as $\mathbf{X}^{\mathbf{E}} \in \mathbb{R}^{T \times C}$, where $C$ is the number of EMG channels with $T$ discrete time steps. The EMG signals undergo standard minimal preprocessing, similar to past works~\cite{gaddy-klein-2021-improved, gaddy2020digital}. Since the original sampling rate is high (>800 Hz), we leverage a temporal 1D convolutional layer with a stride of $N$ ($N=6$ in our case) to facilitate downsampling to $T/6$. Similar to~\citet{gaddy-klein-2021-improved}, we use residual blocks with 1D convolutional layers to extract EMG features, employing a stack of two residual blocks. Next, we explicitly facilitate the learning of sequential dependencies in the extracted features, finding that a bidirectional long-short-term memory (BiLSTM) network effectively captures complex temporal dependencies (further design choices for this sequential block are compared in Section~\ref{subsec:ablate}). This is followed by another 1D convolutional layer with stride $N=2$ for further downsampling, resulting in the embedding $\mathbf{\Tilde{E}} \in \mathbb{R}^{(T/48) \times \Tilde{F}}$, which is then projected using fully connected linear layers to match the input embedding dimension of the LLM, generating the EMG embeddings $\mathbf{E} \in \mathbb{R}^{\Hat{T} \times F}$, where $T > \Hat{T} \approx T/48$ and $F \in \{4096\footnote{\href{https://huggingface.co/meta-llama/Llama-2-7b}{LLaMA 2-7B}}, 3072\footnote{\href{https://huggingface.co/meta-llama/Llama-3.2-3B}{LLaMA 3.2-3B}}\}$ in our case. We use GeLU activation function~\cite{baevski2020wav2vec}.

\begin{figure}
    \centering
    \includegraphics[width=\linewidth]{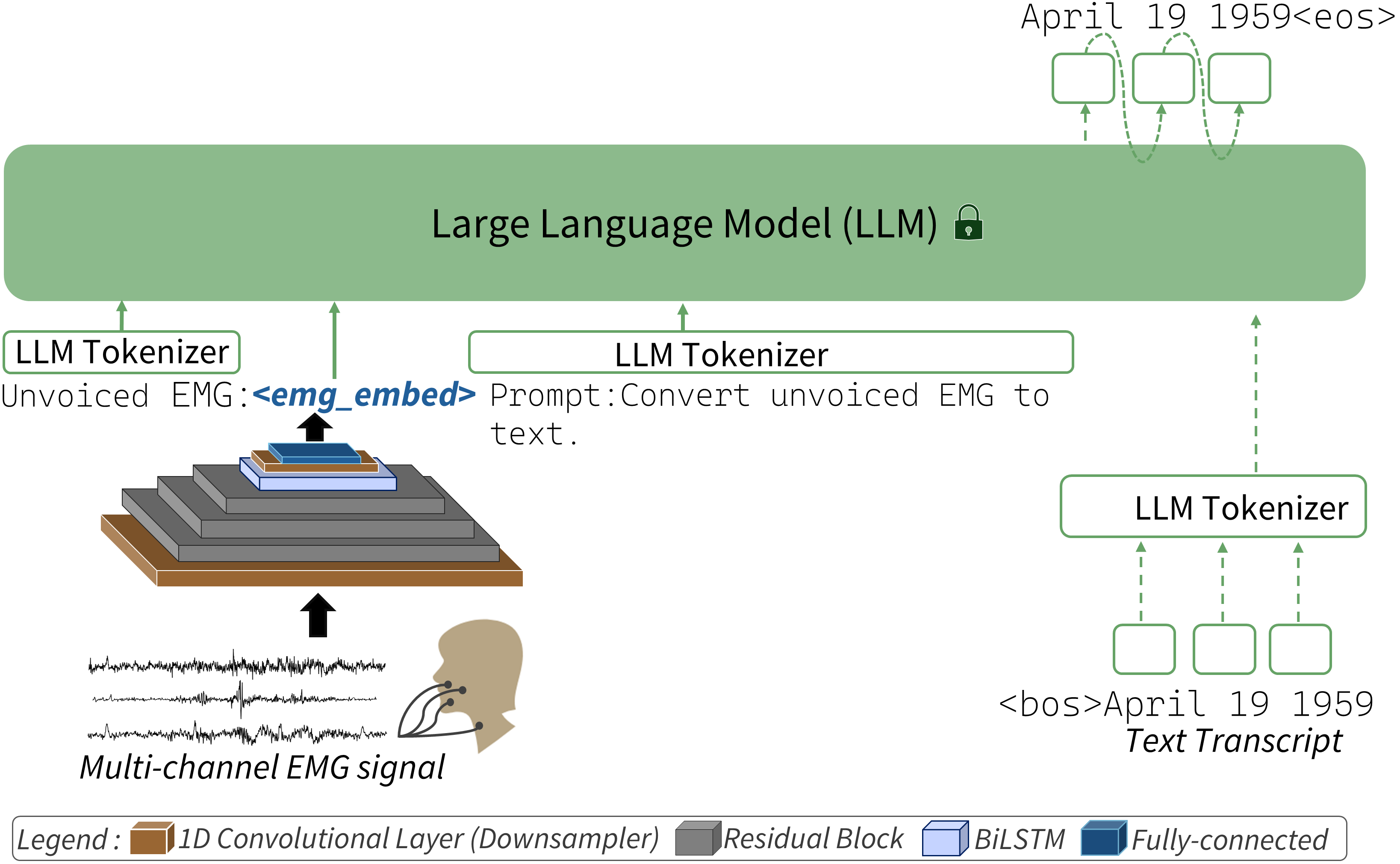}
    \caption{Our trainable EMG adaptor with frozen LLMs to transcribe text from only unvoiced EMG.}
    \label{fig:overall_arch}
\end{figure}

\smallskip
\noindent \textbf{Data Preparation for Large Language Models.} Our EMG adaptor network is defined as $\mathcal{G}: \mathbf{X}^{\mathbf{E}} \rightarrow \mathbf{E}$. We contextualize the embeddings by prepending them with a text identifier $\mathbf{P_1}=\texttt{Unvoiced EMG:}$, and appending a prompt describing the task as $\mathbf{P_2}$=\texttt{Prompt: Convert unvoiced EMG embeddings to text.} The LLM tokenizer converts the text identifier and prompt into text embeddings using the mapping $\mathcal{M}:\mathbf{X}^{\mathbf{P}} \rightarrow\mathbf{H}^{\mathbf{P}}$, where $\mathbf{X}^{\mathbf{P}}=[\mathbf{P_1}, \mathbf{P_2}]$. To prepare the input for the LLM, the embeddings from the prompts are concatenated with the EMG embeddings. 

\smallskip
\noindent \textbf{Training Framework.} For each unvoiced EMG signal, we have a corresponding text transcription \( \mathbf{X}^{\mathbf{S}} \). Following the recommendations of \citet{gaddy-klein-2021-improved}, we simplify our target by removing punctuation and converting all text to lowercase. We extract embeddings from the final LLM layer and compute the predicted logits \( z_{t_s,y'} \) for each vocabulary item \( y' \) at position \( t_s \). The cross-entropy loss over time steps is given by:
\vspace{-5pt}
\[
\mathcal{L} = - \sum_{t_s=1}^{T_s} \sum_{y' \in \mathcal{V}} y'_{t_s} \log \frac{\exp(z_{t_s,y'} / \tau)}{\sum_{v \in \mathcal{V}} \exp(z_{t_s,v} / \tau)},
\] 
where \( \tau = 0.8 \) is the temperature parameter and \( y'_{t_s} \) is the true class at \( t_s \).
We follow standard recommendations for fine-tuning LLMs, employing the AdamW optimizer~\cite{loshchilov2019decoupled} with a maximum learning rate of \( 5 \times 10^{-5} \) and weight decay. During inference, we autoregressively generate~\cite{dubey2024llama} the predicted target sequence with a beam-width~\cite{freitag2017beam} of 4. More implementation details are provided in Appendix~\ref{app:train_detail}. Our codebase is release 

\section{Experimental Results and Discussion}\label{sec:eval}
\noindent \textbf{Datasets.} We primarily used the single-speaker, 8-channel, closed-vocabulary dataset from~\citet{gaddy-klein-2021-improved}, which comprises 67 words and approximately 26 minutes of unvoiced EMG data across 500 utterances. More details are provided in Appendix~\ref{app:dataset}. We accessed only the unvoiced EMG modality.


\smallskip
\noindent \textbf{Baselines and Experimental Setup.} We use the transducer model proposed by \citet{gaddy-klein-2021-improved} as an application-specific baseline. Our analysis leverages two LLMs: Llama2-7B and Llama3-3B. We also fine-tune Llama3-3B using low-rank adaptation (LoRA)~\cite{hu2022lowrank}, training 0.13\% of its parameters. We use three-fold validation and report word error rate (WER) statistics. Following standard practice, data is split 8:1:1 into training, validation, and test sets. All baselines are trained solely on unvoiced EMG. A minimal code implementation and sample predictions are included in the supplementary materials, with further details in Appendix~\ref{app:train_detail}.

\subsection{Key Findings}
\noindent \textbf{LLMs boost closed-vocabulary unvoiced EMG-to-text conversion by 30\% with minimal data processing.} As shown in Table~\ref{tab:main-results}, compared to the application-specific (App-Specific) model with transformers (54M trainable parameters), the proposed EMG adaptor (EMG-Ad) with frozen LLMs, using only 6M trainable parameters, achieves a 0.52 WER with raw EMG signals as input, outperforming the App-Specific model's 0.75 average WER. While the closed-vocabulary setting poses a challenge due to limited training data, it supports the hypothesis that LLMs, through large-scale training, have likely learned universal language representations that help understand unvoiced EMG with limited datasets. Fine-tuning offered only a 17\% improvement over the App-Specific baseline in this setting, possibly due to overparameterization. 

\smallskip
\noindent \textbf{Handcrafted EMG features improve LLM performance for unvoiced EMG-to-text conversion in closed vocabulary.} Leveraging recommendations from previous works on temporal~\citep{jou2006towards} and spectral~\citep{gaddy2020digital} features, we extract 112 time-varying EMG features and use them as inputs to the baselines. As shown in Table~\ref{tab:main-results}, these handcrafted features consistently outperform raw features across both LLMs used as inputs to the EMG adaptor, showing an average improvement of 15\%. However, for the App-Specific baseline, raw features perform better, similar to~\citet{gaddy-klein-2021-improved}'s findings.

\begin{table}[t]
\caption{Comparison of App-Specific models and EMG adaptors (EMG-Ad) with frozen and fine-tuned LLMs. Frozen parts are shown in \green{\textit{green}}, and the best performance in each setting is in \textbf{bold}. Lower WER is better.}\label{tab:main-results}
\centering
\renewcommand{\arraystretch}{0.9} 
\setlength{\tabcolsep}{4pt}
\footnotesize 
\begin{tabular}{p{0.9cm} l c} 
\toprule
\textbf{} & \textbf{Model} & \textbf{WER} \\
\midrule
\multirow{4}{=}{\raggedright Raw EMG} 
    & App-Specific\tiny\cite{gaddy-klein-2021-improved} & 0.75 $\pm$ 0.06 \\
    & EMG-Ad + \green{\textit{Llama2-7B}} & 0.65 $\pm$ 0.01 \\
    & EMG-Ad + \green{\textit{Llama3-3B}} & \textbf{0.52 $\pm$ 0.05} \\
    & EMG-Ad + Fine-tuned Llama3-3B & 0.62 $\pm$ 0.04 \\
\midrule
\multirow{4}{=}{\raggedright Hand-crafted Features} 
    & App-Specific\tiny\cite{gaddy-klein-2021-improved} & 0.84 $\pm$ 0.06 \\
    & EMG-Ad + \green{\textit{Llama2-7B}} & \textbf{0.49 $\pm$ 0.06} \\
    & EMG-Ad + \green{\textit{Llama3-3B}} & \textbf{0.49 $\pm$ 0.04} \\
    & EMG-Ad + Fine-tuned Llama3-3B & 0.55 $\pm$ 0.02 \\
\bottomrule
\end{tabular}
\end{table}

\begin{figure}[!htbp]
    \centering
    \includegraphics[width=0.9\linewidth]{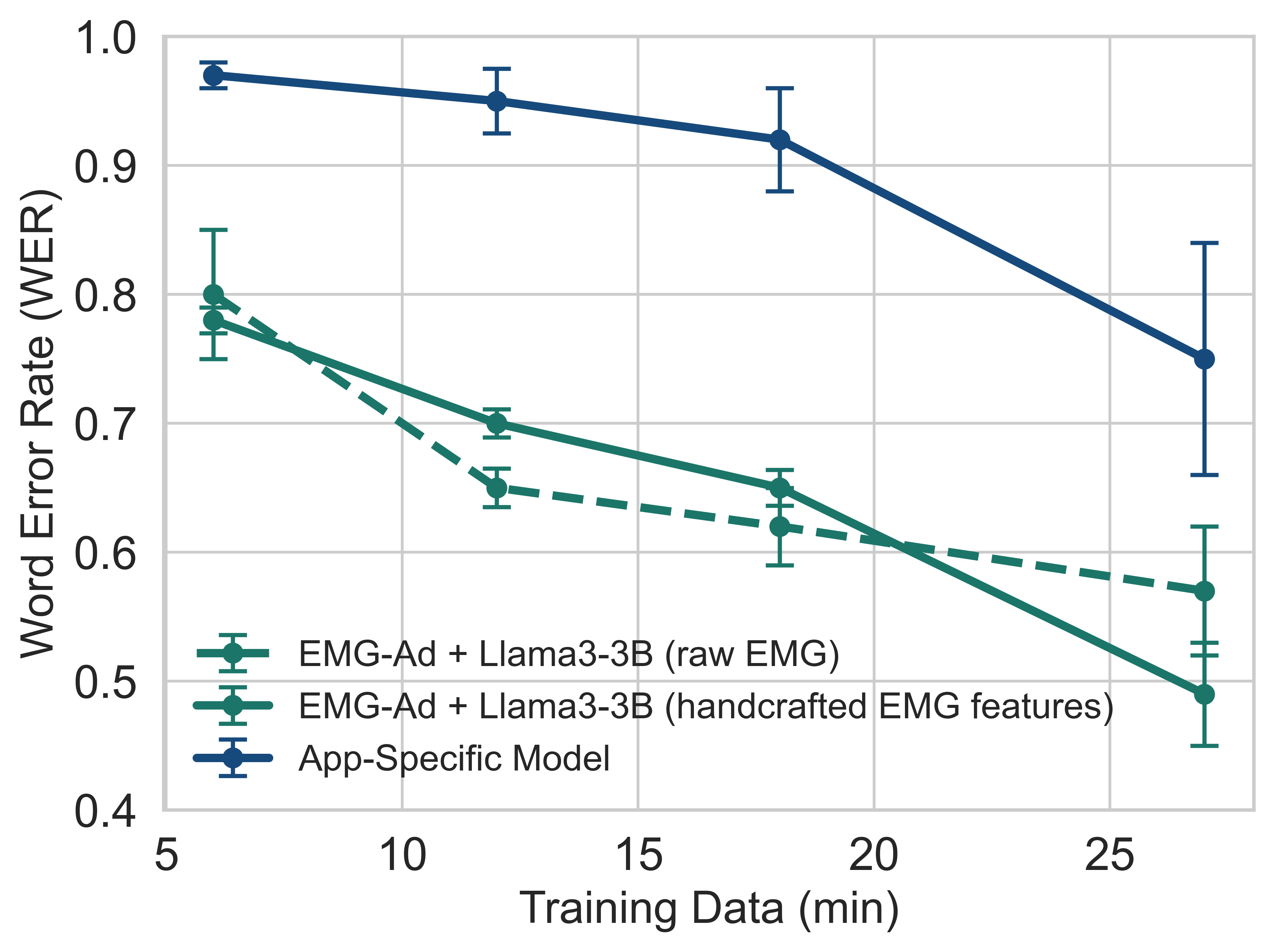}
    \caption{Performance of EMG adaptor with Llama3-3B model vs. App-Specific model across training dataset sizes for unvoiced raw EMG-to-text conversion.}
    \label{fig:data-eff}
\end{figure}

\smallskip
\noindent \textbf{LLMs enable data-efficient learning for closed-vocabulary silent speech.} To further evaluate the effectiveness of LLMs in facilitating learning from a limited number of samples, we randomly subsampled the training data from approximately 26 minutes to 6 minutes, as illustrated in Figure~\ref{fig:data-eff}. Although the WER increases with the reduced training set, our LLM-based approach still outperforms the App-Specific baseline by an average of 26\%. Evidenced by prior work~\cite{diener20_interspeech} and our pilot study in Section~\ref{subsec:further}, surface-EMG signals exhibit distinct person-specific phenotypes. Thus, learning from a limited of samples facilitates building personalized silent-speech interfacing models for LLM assistants.

\begin{figure}[!htbp]
    \centering
    \vspace{-5pt}
    \includegraphics[width=\linewidth]{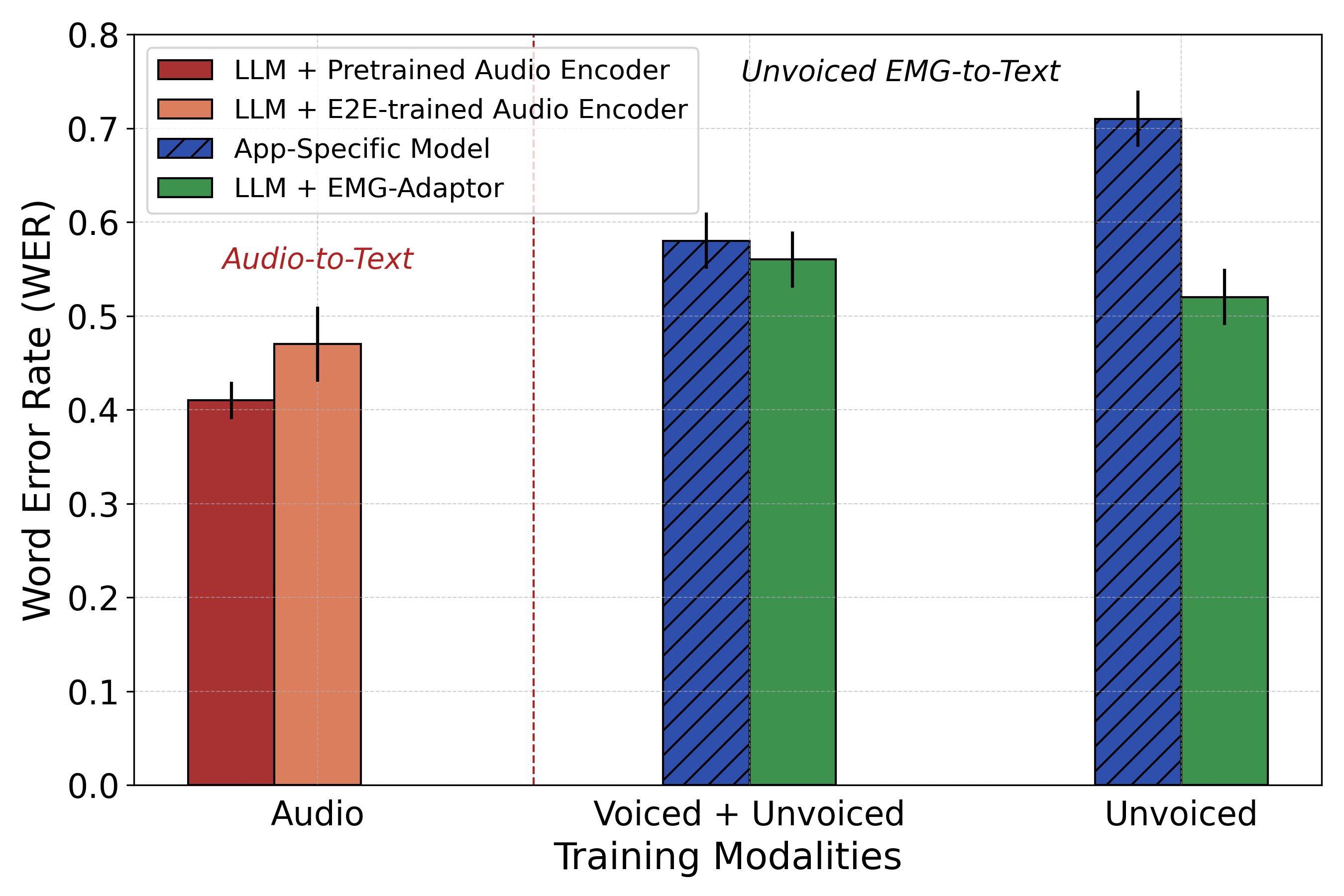}
    \vspace{-20pt}
    \caption{Performance comparison in (1) expanding LLMs to the audio vs. EMG modality, and in (2) training our LLM-based approach and the specialized model using voiced vs. unvoiced EMG data.}
    \label{fig:modality_ood}
    \vspace{-15pt}
\end{figure}

\smallskip
\noindent \textbf{Expanding LLMs to EMG is harder than audio.} To demonstrate the potential of additional language modalities in expanding text-based LLMs, we adopt~\citet{ma2024embarrassingly}'s strategy of incorporating a speech encoder—both an end-to-end trained speech encoder using mel-frequency spectrogram features and a pretrained encoder~\citep{baevski2020wav2vec}—with a linear-projection head into an LLM to convert speech to text (more details in Appendix~\ref{app:audio}). LLMs learn 33\% better from audio even with this simple approach, highlighting the overall task complexity of unvoiced EMG-to-text, as shown in Figure~\ref{fig:modality_ood}.

\smallskip
\noindent \textbf{Augmenting training dataset with voiced EMG benefits specialized models more than LLMs.} In an augmented setting, where we trained using voiced and unvoiced raw EMG, the additional modality led to a 20\% improvement in the App-Specific model (expected behavior as~\citet{gaddy-klein-2021-improved}), while it offered little benefit to the LLM-based approach in this closed vocabulary setting as shown in Figure~\ref{fig:modality_ood}. This suggests that more dedicated efforts in instruction tuning or explicit pairing of voiced and unvoiced EMG~\cite{xu2022multiinstruct} may be needed to learn improved representations from LLMs. However, in this paper, our focus remains on converting unvoiced EMG to text.

\subsection{Ablation Analysis}\label{subsec:ablate}
Table~\ref{tab:ablation} summarizes key variants of the EMG adaptor network design with raw unvoiced EMG, omitting downsampling via high-stride 1D CNN layers. Unlike specialized EMG-to-text models~\cite{gaddy-klein-2021-improved} that benefit from transformers, we find LSTMs perform better in this setting~\cite{mohapatra2023effect}. This may be due to the shorter sequence length (average of four words per utterance with more details in Appendix~\ref{app:dataset}) in the closed-vocabulary dataset and our ability to leverage the language-pretrained transformer layers in LLMs. Appendix~\ref{app:lstm_tf} presents additional ablation results on the sequential backbone architectures.

\begin{table}[!htbp]
    \centering
    \footnotesize
    \renewcommand{\arraystretch}{1.2} 
    \setlength{\tabcolsep}{2.0pt} 
    \caption{Ablation Study of EMG Adaptor training.}
    \begin{tabular}{>{\raggedright\arraybackslash}p{2.0cm} >{\raggedright\arraybackslash}p{4.0cm} c} 
        \toprule
        \textbf{Component} & \textbf{Variants} & \textbf{WER} \\
        \midrule
        \multirow{4}{=}{EMG-Adaptor w/ Llama3-3B} 
        & Fully Connected & 0.70 \\
        & ResBlock (2) & 0.64 \\
        & ResBlock (2) + Transformer & 0.79 \\
        & ResBlock (2) + LSTM & 0.53 \\
        \midrule
        \multirow{2}{=}{Objective w/ Llama2-7B} 
        & Cross-Entropy\tiny(Section~\ref{sec:approach}) & 0.65\\
        & CTC\tiny~\cite{graves2006connectionist} & 0.70 \\
        \bottomrule
    \end{tabular}
    \label{tab:ablation}
    \vspace{-15pt}
\end{table}

Previous works with specialized models~\cite{gaddy-klein-2021-improved, benster2024cross} generally optimize connectionist temporal classification (CTC)~\cite{graves2006connectionist} loss with a high beam width (>100). However, most LLMs are decoder-only architectures and are trained with cross-entropy (CE) loss. One challenge in optimizing these embeddings using CTC loss is ensuring that their temporal length exceeds the target sequence length~\cite{sudo2025joint} for stable optimization. To achieve this, we leverage 1D convolution layers to dilate the embeddings. However, we find that optimizing the embeddings extracted from these LLMs with CTC loss remains suboptimal compared to using CE loss with temperature and small beam widths of just 4 for inference.

%

\subsection{Further Explorations}\label{subsec:further}
\noindent \textbf{Person-identification from unvoiced EMG with 96\% accuracy.} Physiological signals often carry person-specific phenotypes~\cite{zlatintsi2023prevention, mohapatra2023person}. To validate that EMG signals also encode such individualized traits, we conduct a pilot analysis on a public multi-subject dataset~\cite{diener2020cslemgarray}, containing 1,000 unique utterances from four participants. We collapse the LLM's embeddings along the temporal dimension and train a simple classification head (details in Appendix~\ref{app:pid}), achieving an average accuracy of 0.96. This strong performance, consistent with findings in other EMG settings~\cite{diener20_interspeech, wand2009impact}, reaffirms that unvoiced EMG exhibits distinct user-specific patterns—even when users speak the same text segment. To further support this, we also train a fully end-to-end (non-LLM) model for person identification, which achieves 0.99 accuracy, confirming that these signals are highly discriminative. Importantly, our goal is not to propose a state-of-the-art method for user identification using LLMs, but rather to highlight that unvoiced EMG signals inherently carry identifiable traits. This motivates the need for personalized modeling, which typically requires learning from limited data. In this context, our design choice—to use a lightweight trainable adaptor with frozen LLMs—offers a practical and efficient solution. It enables learning user-specific representations with minimal data, a key requirement in personalized EMG-based systems. Our results further emphasize the importance of data-efficient learning, as demonstrated in Figure~\ref{fig:data-eff}.

\smallskip
\noindent \textbf{Exploration of data augmentation for unvoiced EMG.} We explored two data augmentation schemes: (1) random temporal shifts per EMG channel~\cite{gaddy-klein-2021-improved, gaddy2020digital} and (2) Hilbert-transform-based phase augmentation from limb-based EMG gesture recognition~\cite{mohapatra2024phase, wang2025split}, but neither significantly improved performance. Developing tailored augmentation methods for unvoiced EMG—balancing diversity and phonetic coherence—could enhance learning from limited data.


\section*{Conclusion}
Our approach demonstrates the potential of using LLMs to convert unvoiced EMG signals to text, achieving a 0.49 WER without any voiced data. In data-poor settings, it outperforms specialized models by 26\%. Our experiments also highlight the value of hand-crafted features as input to LLMs for this task. This work\footnote{\scriptsize Our processed dataset, codebase and sample predictions are available at \url{https://github.com/payalmohapatra/SilentSpeechLLM}.} helps enable users who cannot produce vocal speech to interact with LLMs through unvoiced commands, especially as LLM-based assistants become ubiquitous.

\section*{Acknowledgments}
We gratefully acknowledge the support from National Science Foundation grants 2038853, 2324936, and 2328032. We would  like to thank the creators of the open-source dataset used in our study. Special thanks to Neil Zhang for his thoughtful feedback and discussions on language modeling methods.


\section*{Limitations}
\noindent \textbf{Exploration with open vocabulary.} In most LLMs, the heavy lifting of supporting large vocabulary sizes (32,000 in Llama2 and 128,000 in Llama3) is carried out in the final layer~\cite{wijmans2024cut}. This makes learning even a closed vocabulary (67 words, approximately 4 words per utterance, with the potential for 50,000 unique utterances) from a new language modality a challenging task, which is the focus of this paper. Supporting a larger, in-the-wild open vocabulary using unvoiced EMG is even more complex, with its difficulty exacerbated by the overall lack of data and the need for personalized models. However, our current explorations lay the groundwork for exploring this direction next. Prior specialized approaches in this domain heavily rely on the availability of audio and voiced EMG, typically using CTC loss for optimization. This also presents an additional challenge in training LLMs in low-resource modalities with a new objective that does not effectively utilize their learned embeddings from large-scale data, leading to suboptimal training. These challenges can be addressed by exploring multimodal LLMs to leverage implicitly aligned embeddings or by reformulating the open vocabulary as a larger closed set and using target-steering methods to train on this restructured vocabulary~\cite{han2024word}.

\smallskip
\noindent \textbf{Broader Extension to more datasets and languages.} Specialized EMG-to-text models are often designed for a predetermined EMG configuration or are deeply tied to phonetic auxiliary tasks, limiting them to the English language and a specific dataset. Our method has the potential to be extended to multilingual and multi-configuration instrumentation. However, due to the lack of publicly available closed-vocabulary datasets in such settings, we limit our investigation to English corpora. A systematic multilingual and diverse instrumentation closed- and open-set recording setup can accelerate exploration in this challenging direction of converting unvoiced EMG to text. 

Additionally, our current approach is reliant on the embedding layers of LLMs, so it cannot work with language model APIs that do not provide direct access to these embeddings.

\section*{Ethical Concerns and Potential Risks} 
In this work, we utilize pretrained LLMs, specifically Llama, in accordance with their usage license, solely for academic research purposes. We do not foresee any immediate ethical concerns arising from our work. However, as an LLM application for interpreting biosignals, appropriate measures must be taken to preserve user privacy. Our techniques help make LLM-based assistants accessible to speech-impaired users, thereby encouraging socially beneficial outcomes.


\bibliography{acl_main}

\appendix

\newpage
\appendix{APPENDIX}
\section{Dataset Statistics}\label{app:dataset}
\begin{figure}[!htbp]
    \centering
    \includegraphics[width=\linewidth]{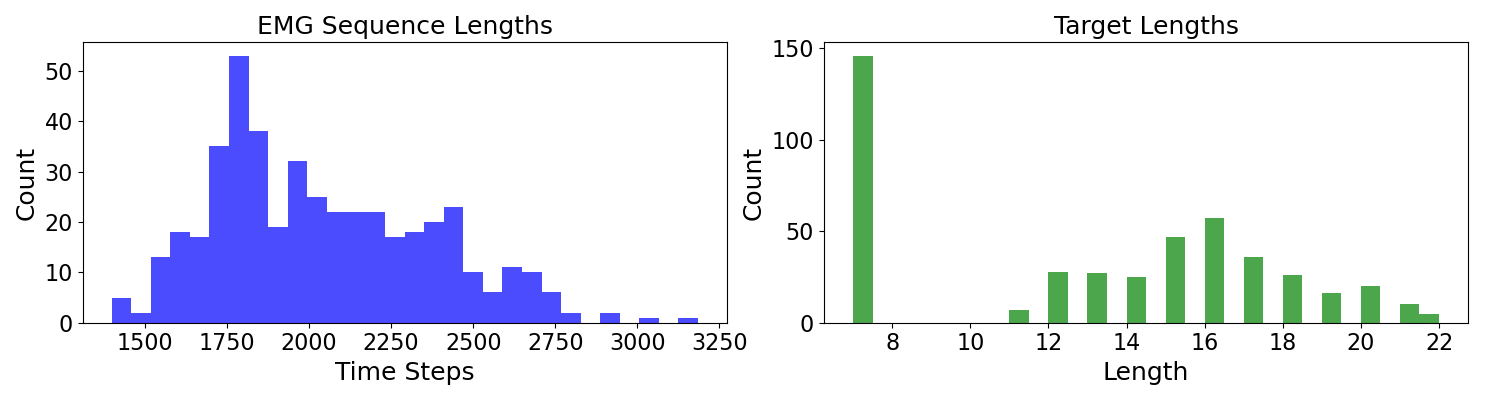}
    \caption{Histogram of the distribution of the sequence lengths and the target lengths for~\citet{gaddy2020digital} closed-vocabulary dataset.}
    \label{fig:gaddy_stat}
\end{figure}

\begin{figure}[!htbp]
    \centering
    \includegraphics[width=\linewidth]{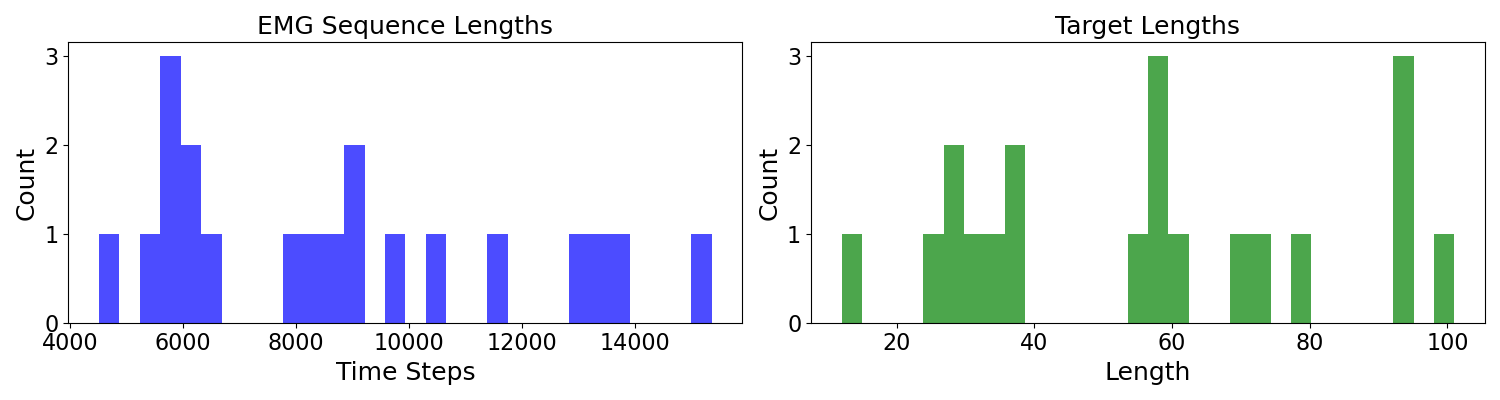}
    \caption{Histogram of the distribution of the sequence lengths and the target lengths for~\citet{diener2020cslemgarray}'s Person 1 Block1 Initial segment of the data which is the superset for all the utterances.}
    \label{fig:csl_stat}
\end{figure}

Table~\ref{tab:gaddy_csl} provides a summary of the dataset statistics, while Figures~\ref{fig:gaddy_stat} and~\ref{fig:csl_stat} illustrate the distribution of the input EMG sequence lengths and target lengths.

\begin{table}[!htbp]
    \centering
    \footnotesize  
    \setlength{\tabcolsep}{3pt}  
    \renewcommand{\arraystretch}{1.1}  
    \caption{Comparison of dataset statistics from~\citet{gaddy2020digital} and~\citet{diener2020cslemgarray}. Here, \( N \) represents the number of participants, and "Train" includes both training and validation samples.}
    \resizebox{\columnwidth}{!}{
    \begin{tabular}{p{3.8cm} p{2.4cm} p{2.4cm}}
        \toprule
        \textbf{Statistic} & \textbf{Gaddy} & \textbf{CSL} \\
        \midrule
        Dataset Size & 500 utterances & 1000 utterances \\
        Target Type & Closed Vocabulary Template & Fixed Open Vocabulary \\
        Number of Individuals (\( N \)) & 1 & 4 \\
        Train Set (Train + Val) & 450 & 800 \\
        Evaluation Set & 50 & 200 \\
        k-fold Evaluations & 3 & 3 \\
        Sampling Rate & 1000 Hz $\rightarrow$ 800 Hz & 2048 Hz \\
        \bottomrule
    \end{tabular}}
    \label{tab:gaddy_csl}
\end{table}

\section{Implementation Details}\label{app:train_detail}

We provide more implementation-specific details of the baselines here and summarize the sizes of all models in Table~\ref{tab:model_parameters}.

\begin{table}[!htbp]
    \centering
    \footnotesize
    \setlength{\tabcolsep}{3pt} 
    \renewcommand{\arraystretch}{1.1}  
    \caption{Trainable parameters of different models. Values are reported in millions (M) and thousands (k).}
    \resizebox{\columnwidth}{!}{
    \begin{tabular}{p{5.5cm} r}
        \toprule
        \textbf{Model} & \textbf{Trainable Parameters} \\
        \midrule
        Application-Specific Model & 54M \\
        Application-Specific (Modified) & 8.1M \\
        LLaMA-7B + EMG Adaptor, LLaMA-3B + EMG Adaptor & 6.4M \\
        Fine-tuned LLaMA-3B + EMG Adaptor & 10.3M \\
        Fine-tuned LLaMA-3B + EMG Adaptor (Alternative) & 5.94M \\
        LLaMA-3B + Audio Adaptor & 590k \\
        \bottomrule
    \end{tabular}}
    \label{tab:model_parameters}
\end{table}

\subsection{Application-specific  Baseline}\label{app:gaddy_baseline}
To test the application-specific baseline model under fair settings, we ran it with beam widths of 4 (same as our LLM-based model) and 100. As noted in Table~\ref{tab:wer_comparison}, the WER for both the beam widths are nearly the same, indicating that beam width has a negligible effect on the prediction in the case of the closed vocabulary. We also train the baseline model with smaller trainable parameters (8M) by reducing the feature size, size of feedforward linear layers, and number of layers in the transformers to 512, 512, and 2 respectively. As shown in Table~\ref{tab:wer_comparison}, the WER with a smaller baseline model is also in the same range as with 54M parameters. This highlights the fact that the difference in performance between our model and the baseline model in Table~\ref{tab:main-results} is due to the choice of LLM for the prediction. 


\begin{table}[!htbp]
    \centering
    \footnotesize
    \setlength{\tabcolsep}{6pt}  
    \renewcommand{\arraystretch}{1.2}  
    \caption{WER performance comparison of the original (54M) and reduced (8M) Gaddy models under different beam-width settings.}
    \begin{tabular}{l l r}
        \toprule
        \textbf{Model} & \textbf{Beam-width} & \textbf{WER} \\
        \midrule
        \multirow{2}{*}{Original (54M)} & \( n = 4 \) & 0.70 \\
                                        & \( n = 100 \) & 0.72 \\
        \midrule
        Reduced (8M) & \( n = 4 \) & 0.87 \\
        \bottomrule
    \end{tabular}
    \label{tab:wer_comparison}
\end{table}

\subsection{Additional Ablation: EMG-Adaptor Backbone Variants}\label{app:lstm_tf}

We conducted additional ablation studies to explore different sequential backbones for the EMG-Adaptor (EMG-Ad) when used with the LLaMA 3B LLM. The results are summarized in Table~\ref{tab:emg_backbone_ablation}.

\begin{table}[ht]
\centering
\scriptsize
\caption{Performance of different sequential backbones for the EMG-Adaptor with LLaMA 3B. Number of transformer layers are denoted as L.}
\label{tab:emg_backbone_ablation}
\setlength{\tabcolsep}{4pt}
\renewcommand{\arraystretch}{1.1}
\begin{tabular}{@{}lcc@{}}
\toprule
\textbf{Sequential Backbone} & \textbf{Params} & \textbf{WER} \\
\midrule
BiLSTM & 5.94M & 0.52 \\
LSTM & 5.5M & 0.58 \\
Transformer (6L + Sinusoidal)~\cite{vaswani2017attention} & 6.7M & 0.79 \\
Transformer (6L + RoPE)~\cite{touvron2023llama} & 6.7M & 0.75 \\
Transformer (2L + RoPE) & 5.3M & 0.72 \\
\bottomrule
\end{tabular}
\end{table}

We experimented with Rotary Position Encoding (RoPE)~\cite{touvron2023llama}, motivated by its effectiveness in the LLaMA models, with the intuition that it might produce better input tokens for the LLM. Although RoPE-based transformers improved performance over the vanilla sinusoidal variant, they still underperformed compared to LSTM-based models.

This aligns with findings from prior time-series literature. For example,~\citet{zeng2023transformers} have argued that the permutation-invariant nature of self-attention leads to temporal information loss. Similarly, practical studies such as IMUPoser~\cite{mollyn2023imuposer} empirically corroborate that LSTMs outperform transformers in fine-grained time-series tasks like human activity recognition, e.g., stating in Section 4.1 of ~\cite{mollyn2023imuposer}: \textit{“Although we did experiment with newer architectures such as transformers, we found these models did not perform well in practice.”}.

While advanced positional encoding schemes and specialized architectures~\cite{wu2021autoformer, nie2022time} have enhanced transformer performance on time series, we restrict our analysis to vanilla transformers for simplicity and focus on their ability to generate suitable EMG tokens for LLMs. Our findings suggest that LSTMs are better suited for this task. This observation is currently limited to short sequences drawn from a closed vocabulary. Future work will investigate more specialized transformer-based architectures for unvoiced EMG modeling.

\subsection{Expanding LLMs to Audio}\label{app:audio}

In this experiment, we primarily leverage the implementation of~\citet{tang2023salmonn} and the idea proposed by~\citet{ma2024embarrassingly} to employ an embarrassingly simple approach for speech recognition with LLMs, using a linear projector from a frozen speech encoder. We use the wav2vec 2.0~\cite{baevski2020wav2vec, mohapatra2023efficient, mohapatra2024missingness, mohapatra2022speech} BASE architecture as our speech encoder, which produces a 768-dimensional feature vector. This vector is then passed through two linear layers to generate the 3072-dimensional input required for Llama3-3B. While the speech encoder can be replaced with alternatives such as HuBERT~\cite{hsu2021hubert} or Whisper~\cite{radford2023robust}, our goal is not to optimize speech-to-text conversion. Instead, we aim to demonstrate that while both audio and EMG involve expanding an LLM’s capability to a new language modality, integrating EMG signals poses significantly greater challenges.

\subsection{Additional Reproducibility Information}

All experiments were conducted using NVIDIA A100 GPUs (3 available, 40GB CUDA memory) and TITAN RTX GPUs (4 available, 24GB CUDA memory) with a maximum runtime of 12 hours in PyTorch. Hyperparameter tuning combined manual and random search, typically requiring fewer than five runs, with selection based solely on validation loss and WER.

The batch size for LLM experiments was 8, and the maximum number of epochs was 500. For the App-Specific model, we retain the original settings~\cite{gaddy-klein-2021-improved}, where the authors re-batchified the input by rolling the temporal dimensions to support training on longer sequences.


\section{Person Identification using Unvoiced EMG : Implementation Details}\label{app:pid}
For the person identification task, we use LLaMA 3.2-3B. Overall the model architecture remains the same as shown in Figure~\ref{fig:overall_arch} except we do not use any prompt in this case. The logits output from the LLM, $Z\in\mathbb{R}^{T \times \mathcal{V}}$, is reduced to $\overline{Z}\in\mathbb{R}^{1\times\mathcal{V}}$ by taking mean along the time axis. The $\overline{Z}$ is then fed into several linear layers to predict logits $y^p\in\mathbb{R}^{1 \times 4}$ as there are 4 distinct persons in the dataset \cite{diener2020cslemgarray}. Due to the linear layers, the number of trainable parameters is 38M, which is higher than the numbers mentioned in Table~\ref{tab:model_parameters} for LLM-based models. Using $y^p$ and actual person ID, the softmax loss is calculated to fine-tune 32M parameters.

\section{Additional Ethical Statements} 
In preparing this work, we only used AI assistants in the capacity to polish the language in the manuscript.


\end{document}